\def\year{2019}\relax
\documentclass[letterpaper]{article} %DO NOT CHANGE THIS
\usepackage{aaai19}  %Required
\usepackage{times}  %Required
\usepackage{helvet}  %Required
\usepackage{courier}  %Required
\usepackage{url}  %Required
\usepackage{graphicx}  %Required
\usepackage{booktabs}
\usepackage{multirow}
\usepackage{textcomp}
\usepackage{boxedminipage}
\usepackage{xcolor}
\usepackage{amsmath}
\usepackage{mathrsfs}
\usepackage{microtype}

\nocopyright

\mathchardef\mhyphen="2D

\newcommand{\uline}[1]{\underline{#1}}

\newcommand{\mytextuparrow}{$\uparrow$}
\newcommand{\mytextdownarrow}{$\downarrow$}

\begin{document}
\title{Challenges in the Automatic Analysis of Students' Diagnostic~Reasoning}

\author{Claudia Schulz\textsuperscript{1},
Christian M.~Meyer\textsuperscript{1},
Michael Sailer\textsuperscript{2},
Jan Kiesewetter\textsuperscript{3},
Elisabeth Bauer\textsuperscript{2},
\\ {\bf \Large Frank Fischer\textsuperscript{2}, 
Martin R.~Fischer\textsuperscript{3}, 
Iryna Gurevych\textsuperscript{1}}
\\ \textsuperscript{1}{Ubiquitous Knowledge Processing (UKP) Lab, Technische Universit\"at Darmstadt, Germany}\\
\textsuperscript{2}{Chair of Education and Educational Psychology, LMU M\"unchen, Germany}\\
\textsuperscript{3}{Institute of Medical Education, University Hospital, LMU M\"unchen, Germany}\\
\url{http://famulus-project.de}}

\maketitle

\begin{abstract}
Diagnostic reasoning is a key component of many professions.
To improve students' diagnostic reasoning skills, educational psychologists analyse and give feedback on epistemic activities used by these students while diagnosing, in particular, hypothesis generation, evidence generation, evidence evaluation, and drawing conclusions.
However, this manual analysis is highly time-consuming.
We aim to enable the large-scale adoption of diagnostic reasoning analysis and feedback by automating the epistemic activity identification. We create the first corpus for this task, comprising diagnostic reasoning self-explanations of students from two domains annotated with epistemic activities.
Based on insights from the corpus creation and the task's characteristics, we discuss three challenges for the  automatic identification of epistemic activities using AI methods:  the correct identification of epistemic activity spans, the reliable distinction of similar epistemic activities, and the detection of overlapping epistemic activities. We propose a separate performance metric for each challenge and thus provide an evaluation framework for future research.
Indeed, our evaluation of various state-of-the-art recurrent neural network architectures reveals that current techniques fail to address some of these challenges.
\end{abstract}

%%%%%%%%%%%%%%%%%%%%%%%%%%%%%%%%%%%%%%%%%%%%%%%%%%%%%%%%%%%%%%%%%%
%%%%%%%%%%%%%%%%%%%%%%%%%%%%%%%%%%%%%%%%%%%%%%%%%%%%%%%%%%%%%%%%%%
\section{Introduction}

\begin{figure*}[t]
  \newcommand{\MARK}[2]{\setlength{\fboxsep}{.75pt}\colorbox{#1}{\vphantom{Pg}#2}}
  \newcommand{\TYPEMARK}[1]{\textit{\uline{#1}}}
  \centering
  \begin{boxedminipage}{\textwidth}
    \raggedright
    \MARK{green}{First I wanted to see if the problem was new, so I checked the teacher's observations}.
    \MARK{yellow}{\TYPEMARK{As it was the same back then},}
    \MARK{yellow}{I ruled out a trauma or another dramatic event}.
    \MARK{cyan}{I was then undecided between autism and ADHD}, \TYPEMARK{since his social} \linebreak
    \TYPEMARK{behaviour seems to be problematic and that's a sign for both diagnoses}.
    \MARK{yellow}{In the end, I settled on ADHD  \TYPEMARK{since his script}} \MARK{yellow}{\TYPEMARK{ seems chaotic and unorganised} and \TYPEMARK{because he seems to have some friends despite his difficult behaviour}}.
	\end{boxedminipage}
	\caption{Exemplary diagnostic reasoning text from the teaching domain, annotated with epistemic activity segments: \MARK{green}{evidence generation}, \TYPEMARK{evidence evaluation}, \MARK{yellow}{drawing conclusions}, \MARK{cyan}{hypothesis generation}.}
	\label{fig:reasoning_text}
\end{figure*}

Diagnostic reasoning is a crucial skill in many professions: Physicians determine a patient's disease based on clinical tests, teachers recognise behavioural disorders in children based on observations, and engineers debug errors in machines or programs based on their analyses of log files or sensor data.
To become competent employees in these professions, students thus need to actively engage with their own diagnostic reasoning behaviour to understand the importance, structure, and validity of their reasoning.
However, even if diagnostic reasoning skills are taught as part of the curriculum,
it is rarely possible for instructors to give extensive feedback regarding each student's reasoning process.
Therefore, our vision is
to automatically analyse and evaluate students' diagnostic reasoning by means of innovative educational AI applications, which facilitate teaching and fostering diagnostic skills at large scale.

\citeauthor{FischerEtAl2014} \shortcite{FischerEtAl2014} suggest to capture and analyse diagnostic -- and more generally scientific -- reasoning in terms of eight \emph{epistemic activities}: problem identification, 
questioning, 
hypothesis generation (HG), 
construction and redesign of artefacts, 
evidence generation (EG),
evidence evaluation (EE), 
drawing conclusions (DC), and
communicating and scrutinising.
While educational psychologists can manually identify and analyse these activities to give feedback regarding students' diagnostic reasoning, automatic methods for epistemic activity analysis are required to scale this up. 

We tackle this task by creating the first cross-domain corpus of epistemic activities in diagnostic reasoning.\footnote{Code and data available at  \url{https://github.com/UKPLab/aaai19-diagnostic-reasoning}}
Our corpus consists of 1,200 anonymised German texts written by medical students and pre-service teachers (henceforth called ``teacher students'') while diagnosing, respectively, the cause of a patient's symptoms and the reason underlying the signs of motivational problems or learning difficulties of a pupil.
We annotated all texts with HG, EG, EE, and DC, the main epistemic activities used in diagnostic reasoning texts \cite{LenzerEtAl2017,GhanemEtAl2016}.
Figure~\ref{fig:reasoning_text} shows an example from the teaching domain: A student first generates evidence, evaluates it, and draws an initial conclusion. This is followed by the generation of a hypothesis about two potential developmental disorders, the evaluation of further evidence, and the drawing of a final conclusion.

Our corpus and task formalization has three characteristics that raise three major \emph{challenges} for AI methods aiming to automatically identify epistemic activities:
(C1)~Epistemic activities are bound to neither token nor sentence level, but may constitute arbitrary spans. 
(C2)~The distinction of similar epistemic activities, such as hypothesis generation and drawing conclusions, is highly challenging.
(C3)~Epistemic activities may be nested or overlapping.
We argue that a mere overall performance measure is not adequate for complex tasks such as epistemic activity identification and propose three challenge-specific metrics, each 
measuring how well an automatic identification method tackles this challenge.
Our evaluation of various state-of-the-art recurrent neural network architectures reveals that current techniques are unable to adequately address these challenges (particularly C2 and C3), which cannot be deduced from an overall performance score.

Our contributions can be summarised as follows:
(1) We create and publish a novel corpus of German diagnostic reasoning texts annotated with epistemic activities in two different domains. 
(2) We identify three key challenges of the epistemic activity identification task and propose a performance measure for each, providing an evaluation framework for future AI research progress on each challenge. 
(3) We compare multiple state-of-the-art recurrent neural network architectures for the automatic identification of epistemic activities, showing that some of the challenges of our task are not addressed by current methods.

%%%%%%%%%%%%%%%%%%%%%%%%%%%%%%%%%%%%%%%%%%%%%%%%%%%%%%%%%%%%%%%%%%
%%%%%%%%%%%%%%%%%%%%%%%%%%%%%%%%%%%%%%%%%%%%%%%%%%%%%%%%%%%%%%%%%%
\section{Related Work}

Natural Language Processing approaches have been successfully applied to clinical diagnosis and decision-making \cite{DemnerCM2009,PrakashEtAl2017}.
These works focus on predicting a diagnosis from clinical narratives (i.e.~descriptions of clinical findings) without considering the diagnostic reasoning process. In contrast, 
our goal is to form the basis for fostering students' diagnostic reasoning skills by providing a resource for the automatic identification of students' reasoning steps.

\paragraph{Diagnostic Reasoning.}
 The process of diagnostic reasoning, in particular in terms of epistemic activities \cite{FischerEtAl2014}, has been investigated in both educational and professional contexts \cite{GhanemEtAl2016,LenzerEtAl2017}.
 These studies focus on 
 protocols to capture reasoning while diagnosing. In contrast, we rely on students' self-explanations of their reasoning.
 Furthermore, we annotate epistemic activities as text segments that can have any length rather than using pre-segmented phrases.
Based on datasets from these previous studies, \citeauthor{CsanadiEtAl2016} \shortcite{CsanadiEtAl2016} and \citeauthor{DaxenbergerCGG2017} \shortcite{DaxenbergerCGG2017} automatically classify the pre-segmented phrases into epistemic activities using a conditional random field. \citeauthor{LernerEtAl2016}  \shortcite{LernerEtAl2016} further propose visualisation tools for scientists, illustrating the results of such automatic classification algorithms. Since no existing datasets are publicly available, it is hard to reproduce and build upon these previous efforts. We thus make our corpus and the corresponding analysis software publicly available.

\citeauthor{Meng18} \shortcite{Meng18} propose an annotation schema for problem-solving dialogues. While some of their categories are similar to epistemic activities (e.g.~\emph{analysis}), the schema focuses on dialogues and is limited to their specific robotics use case (e.g.~\emph{query robot}, \emph{algorithmic thinking-variable}). Similar to our work, they test LSTM-based tagging approaches, but neither their implementation nor their data is publicly available.

\paragraph{Argumentation Mining.}
Related to the study of diagnostic reasoning is argumentative reasoning,
which has recently received growing attention from the NLP community.
The focus has been on identifying argument components \cite{LippiT2015a,SchulzEDKG2018}
or whole arguments, made of components (such as premises and claims) as well as attacking and supporting relations between them \cite{MeniniCSV2018,HabernalG2017}.
Like us, \citeauthor{StabG2014a} \shortcite{StabG2014a} and 
\citeauthor{NguyenL2018} \shortcite{NguyenL2018}
investigate arguments in an educational setting by automatically identifying arguments in students' persuasive essays.
These and other works in argumentation mining generally investigate arguments as the \emph{product} (see e.g.~\citeauthor{HabernalG2017} \citeyear{HabernalG2017}) of the argumentative reasoning process. 
In contrast, we analyse the argumentative, in this case diagnostic, reasoning \emph{process}.
Our work thus adds not only a new corpus, but also a new research angle to the field of argumentation mining.

%%%%%%%%%%%%%%%%%%%%%%%%%%%%%%%%%%%%%%%%%%%%%%%%%%%%%%%%%%%%%%%%%%
%%%%%%%%%%%%%%%%%%%%%%%%%%%%%%%%%%%%%%%%%%%%%%%%%%%%%%%%%%%%%%%%%%
\section{A Corpus of Epistemic Activities}

To simulate professional diagnostic decision-making, \emph{case scenarios} that ask students for a diagnosis in a specific problem-solving scenario are frequently used \cite{ThistlethwaiteEtAl2012}. 
The underlying data of our corpus was collected by educational psychologists using eight case scenarios for the medicine and eight for the teaching domain.
In the medicine domain, the anamnesis and results of various medical tests of fictional patients were given to medical students. 
In the teaching domain, the information provided to teacher students comprised observations about fictional pupils' behaviour, a report of grades, and a transcript of a meeting with the parents. 

Diagnostic reasoning of medical and teacher students was captured by means of self-explanations,
prompting them to reflect on their reasoning process \cite{Renkl2014}. 
In each case scenario, the students were asked `What is the diagnosis?' and `How did you come up with this diagnosis?'. The written explanations constitute the \emph{(diagnostic) reasoning texts} used for our corpus. 
Figure~\ref{fig:reasoning_text} depicts such a reasoning text.

\begin{table*}[t]
\small
\centering
 \begin{tabular}{lccccccccc}
  \toprule
  Domain  & $\alpha_U$ & $\alpha_U$-HG & $\alpha_U$-EG & $\alpha_U$-EE & $\alpha_U$-DC & $\alpha_U$-segment & \mytextuparrow $\alpha_U$-pair & \mytextdownarrow $\alpha_U$-pair\\
  \midrule
  medicine & 0.67 & 0.60 &  0.65 & 0.75 & 0.56 & 0.86  & 0.71 & 0.62\\
  teaching & 0.65 & 0.43 & 0.56 & 0.75 & 0.49 & 0.82 & 0.67 & 0.63 \\
  \bottomrule
 \end{tabular}
 \caption{Inter annotator agreement (IAA) in terms of Krippendorff's $\alpha_U$.}
 \label{tab:iaa}
\end{table*}

\begin{table*}[t]
\small
\centering
 \begin{tabular}{lcccccc}
  \toprule
  Domain  &  $\alpha_U$-HG\&DC & $\alpha_U$-EE\&DC &$\alpha_U$-HG\&EE & $\alpha_U$-EG\&EE &$\alpha_U$-EG\&HG &  $\alpha_U$-EG\&DC \\
  \midrule
  medicine & \textbf{0.71} & \textbf{0.85} &\textbf{0.78} &\textbf{0.78} &  0.61 &  0.56\\
  teaching &  \textbf{0.62} & \textbf{0.81} &\textbf{0.77} & 0.72 & 0.47 &  0.48\\
  \bottomrule
 \end{tabular}
 \caption{IAA ($\alpha_U$) when merging epistemic activities. Bold indicates a value higher than both single activities.}
 \label{tab:iaa_merge}
\end{table*}

%%%%%%%%%%%%%%%%%%%%%%%%%%%%%%%%%%%%%%%%%%%%%%%%%%%%%%%%%%%%%%%%%%
\subsection{Guideline Development and Annotation Setup}
To analyse students' diagnostic reasoning, educational psychologists \cite{FischerEtAl2014} suggest a taxonomy of \emph{epistemic activities}.
The ones relevant for our context are:
\emph{hypothesis generation} (HG; the derivation of possible answers to the problem), \emph{evidence generation} (EG; the derivation of evidence, e.g.~through deductive reasoning or observing phenomena),   \emph{evidence evaluation} (EE; the assessment of whether and to which degree evidence supports an answer to the problem), and \emph{drawing conclusions} (DC; the aggregation and weighing of evidence and knowledge to derive a final answer to the problem).

Drawing upon the expertise of educational psychologists and implementing their requirements as to what constitutes an epistemic activity in a reasoning text, we aim to capture epistemic activities as accurately and naturally as possible. Thus, annotators simultaneously 
identify epistemic activity segments and their type (HG, EG, EE, or DC). 
Epistemic activities can therefore have any length and are not restricted to predefined segments.
Our only rule for identifying epistemic activity segments is that any word indicative of the role an epistemic activity plays in the reasoning process should be part of the segment.
Note that the unrestricted segment length also allows for the annotation of overlapping or nested epistemic activity segments.
Especially the definition of DC indicates potential overlaps between DC and EE.

In four pilot annotations with educational psychologists, we further refined the definitions of EG and EE to the context of diagnostic reasoning texts: Due to the case scenario setup, students cannot generate  evidence in the original sense, for example by performing tests or studies, since such evidence is already given in the case scenario information.
We thus define EG as explicit statements 
of obtaining evidence from the case information or of recalling own knowledge,
as shown in Figure~\ref{fig:reasoning_text}.
Moreover, many students do not explicitly evaluate evidence concerning its degree of relevance in supporting or refuting a potential answer, as defined for EE. 
We thus interpret the mentioning of evidence as an active selection of information considered relevant and define EE in this manner.

The resulting cross-domain annotation guidelines are understandable by annotators without previous knowledge about epistemic activities, which makes the guidelines easily adaptable to new domains.
We recruited four expert annotators from the teaching and five from the medicine domain, who are all native German speakers.

Using the open-source INCEpTION Annotation Tool \cite{tubiblio106270}, 
we performed three rounds of annotations, each followed by a discussion to identify and resolve difficulties.
This ensured a high quality of our corpus.
In the first and second annotation rounds, a set of (respectively) 100 and 50 reasoning texts was annotated by all domain annotators.
Since we found satisfactory inter-annotator agreement, each annotator then annotated a different set of 100 texts in the third round.
We thus obtained annotations for 550 reasoning texts in the teaching and 650 in the medicine domain.

%%%%%%%%%%%%%%%%%%%%%%%%%%%%%%%%%%%%%%%%%%%%%%%%%%%%%%%%%%%%%%%%%%
\subsection{Reliability of Annotations}
\label{sec:agreement}
We evaluate the quality of annotations in terms of agreement between the domain annotators in the first two annotation rounds. Since our annotation task involves not only classifying the type of epistemic activity but also identifying segments, we apply Krippendorff's $\alpha_{U}$ \cite{Krippendorff1995} as implemented in 
DKPro Agreement \cite{MeyerMSG2014}.
%instead of commonly used $\kappa$ metrics, which are limited to classification tasks with pre-segmented units.
%The scale of $\alpha_{U}$ is, however, comparable to Krippendorff's $\alpha$ and highly similar to $\kappa$ \cite{Krippendorff04a}.

The first five columns in Table~\ref{tab:iaa} show the overall agreement and the agreement for each epistemic activity.
We note that the agreement between annotators in the medicine domain is slightly higher than in the teaching domain.
This result is in line with comments by the teaching annotators, that reasoning texts vary widely in terms of writing style and terminology,
making the annotation task more difficult than in the medicine domain, where most students use a similar terminology. 
This is further reflected in the lower agreement on segments (when not considering the type of epistemic activity) in the teaching domain ($\alpha_{U}$-segment in Table~\ref{tab:iaa}).
Overall, the agreement of more than 0.8 on segments shows that epistemic activity segments can be reliably identified by human annotators.
Note that the fact that there were five annotators in the medicine and only four in the teaching domain does not affect the comparability of agreement scores:
when forming groups of four medicine annotators, the $\alpha_U$ scores are 0.66, 0.66, 0.66, 0.67, and 0.68,
which are in correspondence with the score of all five annotators.

To make sure that none of the annotators produced unreliable annotations, we 
compute pairwise agreement scores. The highest and lowest scores are reported in the last two columns of Table~\ref{tab:iaa}, showing that 
even the lowest pairwise agreement still indicates reliable annotations.
Upon closer inspection, none of the annotators has consistently low agreement with all other annotators. 
We thus conclude that all of our annotators produce annotations of similar reliability, justifying the third annotation round with only one annotator per reasoning text.

\begin{table*}[ht]
 \centering
 \small
 \begin{tabular}{l l cccccccccc}
  \toprule
  && EG & EE & HG & DC & EG/EE & HG/DC & DC/EE & EG/HG & HG/EE & EG/DC \\
   \midrule
  \parbox[t]{2mm}{\multirow{3}{*}{\rotatebox[origin=c]{90}{\textbf{MeD}}}}
&\# & 219 & 2124 & 623 & 493 & 5 & 4 & 342 & 0 & 12 & 4\\
&av. \# & 0.35 & 3.27 & 0.96 & 0.76 & -- & --&  --& --& --& --\\
&av len. & 10.1 & 11.6 & 9.0 & 16.0 & 3.8 & 8.5 & 9.8 & -- & 5.7 & 6.8\\
\midrule
  \parbox[t]{2mm}{\multirow{3}{*}{\rotatebox[origin=c]{90}{\textbf{TeD}}}}
& \# & 354 & 2671 & 311 & 444 & 8 & 2 & 143 & 3 & 8 & 3\\
&av. \# & 0.64 & 4.86 & 0.57 & 0.81 & -- & --&  --& --& --& --\\
&av. len. & 12.4 & 12.1 & 13.5 & 15.4 & 7.9 & 22.0 & 10.9 & 6.0 & 11.1 & 11.7\\
  \bottomrule
 \end{tabular}
\caption{Corpus statistics in terms of absolute number~(\#), average number per text (av. \#), and
average number of tokens (av. len), where EE/EG (and similar) denotes an overlap of an EG and EE segment.}
\label{tab:statistics}
\end{table*}

%%%%%%%%%%%%%%%%%%%%%%%%%%%%%%%%%%%%%%%%%%%%%%%%%%%%%%%%%%%%%%%%%%
\subsection{Annotation Difficulties}
The agreement scores in Table~\ref{tab:iaa} indicate that HG and DC are the most difficult to identify amongst the epistemic activities.  In Table~\ref{tab:iaa_merge}, we further investigate this difficulty and show that the reason for this is that HG and DC are often confused:
when treating HG and DC as a single epistemic activity, the annotators' agreement improves by at least 18\% compared to the agreements for the two separate epistemic activities.
We observe a similar trend when treating EE and DC as a single epistemic activity, where the agreement increases by at least 8\%. This indicates that EE and DC are also difficult to distinguish.
These two distinction problems suggest that it may also be problematic to distinguish HG and EE.
Indeed our findings in Table~\ref{tab:iaa_merge} show that treating the two as a single epistemic activity also increases the agreement.
Overall, our analysis suggests that distinguishing HG, DC, and EE is more difficult in the teaching domain, as we observe a larger increase of agreement when merging the epistemic activities.
This difficulty in distinguishing epistemic activities may prove to be a challenge for AI methods that automatically identify epistemic activities.

EG is seldom confused with other epistemic activities (see Table~\ref{tab:iaa_merge}).
The only distinction problem, especially in the medicine domain, is with EE. 
We attribute this to the implicit use of evidence, such as naming medical tests rather than the results of these tests as evidence.
For example, ``important for the diagnosis are the neurological and physical examinations'' is EE as it provides information considered relevant for making a diagnosis.
However, some annotators marked this as EG, as it implicitly expresses that the student obtained evidence from the case scenario information (namely the neurological and physical examination notes).

%%%%%%%%%%%%%%%%%%%%%%%%%%%%%%%%%%%%%%%%%%%%%%%%%%%%%%%%%%%%%%%%%%
\subsection{Gold Annotations}
\label{sec:goldLabels}
Given our annotations, we generate gold segments and labels for both domains, which results in our final corpus consisting of
550 annotated text from the teaching domain (TeD) and 650 annotated text from the medicine domain (MeD).

\paragraph{Gold Standard Creation.}
For the first and second annotation round, where each text was annotated by all domain annotators,
we first apply majority voting and
subsequently resolve the annotations not decidable in a discussion between the annotators.
In the medicine domain, an annotated segment is considered a gold annotation by majority voting, if this exact segment is annotated as the same epistemic activity by four out of the five annotators.
We opt for a majority of four instead of three to ensure a high quality of our corpus, thus avoiding segments incorrectly labeled by three of the five annotators.
Indeed, 20\% of the 517 annotations not decidable by our majority voting were given a different label during the annotators' discussion than a majority voting of three would have assigned.
In the teaching domain, a segment annotated exactly the same by three out of the four annotators is considered a gold annotation.
For the third annotation round, the annotations of each annotator are taken as gold annotations.

A distinguishing feature of our new corpus is that segments with different labels may \emph{overlap},
i.e.~a token may belong to two different epistemic activity segments.
 Overlapping segments have so far mostly been studied for lower-level tasks such as named-entity recognition \cite{AlexHG2007,Ling2012} or as arising from jointly modeled lower-level tasks \cite{McDonald2005}. However, named entities constitute much shorter segments than the ones found in our corpus. For high-level tasks, overlapping segments provide a new direction of research, which has so far only been explored for discourse units in Wikipedia and news articles \cite{AfantenosDMD2010}, where, in contrast to our corpus, overlapping segments do not belong to different classes.
Due to the lack of resources so far, the identification of overlapping segments may prove to be challenging for AI methods aiming to automatically identify epistemic activities.

\paragraph{Corpus Analysis.}
In an analysis of 
our new corpus, we find that reasoning texts in TeD are over 50\% longer than those in MeD (average token count per text: 100 versus 64).
This difference in length can  be partly attributed to the inclusion of more statements not constituting an epistemic activity by teacher students, on average ten tokens as compared to three tokens by the medical students.
Our analysis also reveals that epistemic activity segments may consist of anything between one or two tokens, and two or three sentences. This supports the requirement of educational psychologists that epistemic activities should be annotated without pre-segmentation.

Table~\ref{tab:statistics} compares the two parts of our corpus in more detail.
We observe that, overall, teacher students use more and longer epistemic activities than medical students.
Interestingly, there is nearly no difference in the length or average number of DCs used in the medicine and teaching domain. 
Furthermore, HG is used more frequently by medical students, while teacher students engage more frequently in EE and EG.
We attribute the former observation to the diagnostic training included in the medical but not in the teacher education curriculum, so medical students are more aware of the importance of forming hypotheses than teacher students.
The latter observation may be due to the less concise writing and reasoning style of teacher students, leading to the annotation of multiple EE segments rather than a single one.
Table~\ref{tab:statistics} also highlights that overlaps between all types of epistemic activities occur in our corpus. As can be expected from the definition of DC, the majority is between DC and EE.

Concluding the analysis of our corpus, we note that
the distribution of epistemic activities is highly skewed, with 62/70\% (MeD/TeD) being EE and only 6/10\% (MeD/TeD) being EG.
This in combination with the fact that annotators found it difficult to distinguish HG, DC, and EE suggests that our new corpus may prove challenging for algorithms aiming to automatically identify epistemic activities.

%%%%%%%%%%%%%%%%%%%%%%%%%%%%%%%%%%%%%%%%%%%%%%%%%%%%%%%%%%%%%%%%%%
%%%%%%%%%%%%%%%%%%%%%%%%%%%%%%%%%%%%%%%%%%%%%%%%%%%%%%%%%%%%%%%%%%
\section{Automatic Identification of Epistemic~Activities}

Having created a high-quality corpus of epistemic activities, which naturally captures the epistemic activity identification task performed by educational psychologists, this section focuses on how to apply machine learning methods to scale up this task.
Our corpus has three intrinsic characteristics: it provides segments of arbitrary length (C1), it distinguishes different epistemic activity types (C2), and it includes overlapping epistemic activity segments (C3).
These can be interpreted as \emph{challenges} that an ideal method should address.

%%%%%%%%%%%%%%%%%%%%%%%%%%%%%%%%%%%%%%%%%%%%%%%%%%%%%%%%%%%%%%%%%%
\subsection{Modeling the Task}
Challenges C1 and C2 imply that we are dealing with a multi-class sequence labeling task, as for example encountered in the related task of argument component identification \cite{SchulzEDKG2018}.
This is commonly modeled by assigning a label to each token that expresses both the type of segment, here the type of epistemic activity $A = \{HG, EG, EE, DC\}$, and the segment boundaries in terms of BIO-labels $S = \{B, I, O\}$, indicating the beginning ($B$), continuation ($I$), or absence ($O$) of a segment. That is, the possible labels for each token are $C = (\{B,I\} \times A) \cup \{O\}$.
However, challenge C3 turns our task into a \emph{multi-label} problem \cite{MultiLabelSurvey}, in which each token may be associated with multiple labels, i.e.~a subset $C' \subseteq C$.

One way to tackle multi-label tasks is through problem transformation. Here, we use three transformation strategies to obtain: 1) multiple (single-label) multi-class problems, 2) a unique  (single-label) multi-class problem, and 3) a multi-dimensional classification problem.

For the first transformation strategy, called \textsc{Separate}, we map the class labels $C$ to four separate class labels $C_{HG} = C_{EG} = C_{EE} = C_{DC} = S$. Each $C_i$ represents a separate multi-class classification problem, i.e.~the identification of each epistemic activity is treated as a separate task. This setup has the obvious drawback of having to optimize each epistemic activity separately.

The second transformation, called \textsc{Concat}, creates a new set of class labels, which consists of all possible combinations of 
different epistemic activity segments
that may be associated with a token, i.e.~concatenations of BIO-labels for each epistemic activity: $C_\mathrm{concat} = S^{|A|}$. For example, the first token of the last sentence in Figure~\ref{fig:reasoning_text} is labeled as $O \mhyphen O \mhyphen O \mhyphen B$ (beginning DC), the following tokens up to `ADHD' as $O \mhyphen O \mhyphen O \mhyphen I$ (continuation DC), `since' as $O \mhyphen O \mhyphen B \mhyphen I$ (beginning EE, continuation DC), all following tokens up to `unorganised' as $O \mhyphen O \mhyphen I \mhyphen I$ (continuation EE and DC), and the full stop as $O \mhyphen O \mhyphen O \mhyphen O$.  The transformation thus results in a single-label multi-class task. In theory, this transformation implies an explosion in possible class labels (namely 81). However, in practice only a small portion of these occur in our data (12 in MeD, 16 in TeD).

Finally, our third transformation, called \textsc{Multi-Output}, models the multi-label task as a multi-output (also called multi-dimensional) classification problem \cite{multiDimensional,MultiOutputSurvey}, where $C_{HG}$, $C_{EG}$, $C_{EE}$, and $C_{DC}$ are the class labels of the four-dimensions to be jointly predicted.

%%%%%%%%%%%%%%%%%%%%%%%%%%%%%%%%%%%%%%%%%%%%%%%%%%%%%%%%%%%%%%%%%%
\subsection{Neural Architectures}
Recurrent Neural Networks are state-of-the-art for multi-class sequence labeling tasks \cite{MaHovy2016}.
We apply \citeauthor{ReimersG2017}'s \shortcite{ReimersG2017} implementation of a bidirectional long short-term memory (BiLSTM) network with a conditional random field (CRF) output layer for the two transformations \textsc{Separate} and \textsc{Concat}, as these result in multi-class tasks.
For \textsc{Separate}, we train a separate BiLSTM-CRF for each epistemic activity, each predicting one of the sequence classes $C_{HG}$, $C_{EG}$, $C_{EE}$, and $C_{DC}$ per token.
For \textsc{Concat}, we train a single BiLSTM-CRF to predict a single $C_\mathrm{concat}$ label per token. 

For \textsc{Multi-Output}, we implement a multi-output architecture similar to \citeauthor{ReimersG2017}'s (\citeyear{ReimersG2017}), but using a shared BiLSTM with multiple CRF output layers instead of only one. Each CRF layer predicts one of the classes $C_{HG}$, $C_{EG}$, $C_{EE}$, and $C_{DC}$.

\begin{table*}[t]
\centering
\small
\begin{tabular}{l l l llll l rlll}
  \toprule
  & & $\mathit{HL}$ &  \multicolumn{4}{c}{$M_S$} &\multicolumn{1}{l}{$M_A$} &
  \multicolumn{4}{c}{$M_O$} 
  \\
  \cmidrule(lr){3-3} \cmidrule(lr){4-7} \cmidrule(lr){8-8} \cmidrule(lr){9-12} 
 & Architecture & all & EG & EE & HG & DC & all & \multicolumn{1}{l}{EG} & EE & HG & DC 
 \\\midrule
   \parbox[t]{2mm}{\multirow{6}{*}{\rotatebox[origin=c]{90}{\textbf{MeD}}}}
  & \textsc{Multi-Output} & 0.07 & 71.60 & 80.20\textsuperscript{+} & 69.28 & 65.32 & 22.21\textsuperscript{+} & 63.09 & 66.39\textsuperscript{+} & 45.50 & 44.76 \\
 & \textsc{Separate} & 0.07 & 70.87 & 80.24\textsuperscript{+} & 68.53 & 65.80 & 21.25\textsuperscript{+} & 63.15 & 65.31\textsuperscript{+} & 50.26 & 49.26 \\
 & \textsc{Concat} & 0.06\textsuperscript{+++} & 71.05 & 79.96\textsuperscript{+} & 69.36 & 65.18 & 23.01\textsuperscript{++} & 67.86 & 66.43\textsuperscript{+} & 44.51 & 45.40 \\\cmidrule(lr){2-12}
 & \textsc{Pref-Baseline} & 0.07 & 70.02 & 75.46 & 69.32 & 65.74 & 19.77 & 52.91 & 38.87 & 46.34 & 49.03 \\
 & \textsc{Maj-Baseline} & 0.11 & 32.70 & 23.49 & 30.48 & 29.96 & \;\;4.25 & 33.13 & 31.00 & 32.61 & 1.39 \\\cmidrule(lr){2-12}
 & human upper bound & 0.04 & 85.61 & 90.25 & 86.37 & 85.58 & 35.06 & 100.00 & 76.15 & 91.38 & 76.50\\ 
\midrule 
\parbox[t]{2mm}{\multirow{6}{*}{\rotatebox[origin=c]{90}{\textbf{TeD}}}}
 & \textsc{Multi-Output} & 0.07 & 78.53 & 78.87\textsuperscript{+} & 57.16 & 61.77 & 19.96\textsuperscript{+} & 58.42 & 71.98\textsuperscript{+} & 32.61\textsuperscript{+} & 47.10 \\
 & \textsc{Separate} & 0.07 & 76.38 & 79.47\textsuperscript{+} & 57.05 & 57.52 & 18.34 & 54.68 & 78.89\textsuperscript{+++} & 32.09 & 36.11 \\
 & \textsc{Concat} & 0.06\textsuperscript{++} & 78.71\textsuperscript{+} & 79.07\textsuperscript{+} & 57.12 & 62.53\textsuperscript{+} & 21.68\textsuperscript{+++} & 56.75 & 68.75\textsuperscript{+} & 32.51 & 51.97\textsuperscript{+} \\ \cmidrule(lr){2-12}
 & \textsc{Pref-Baseline} & 0.06 & 77.60 & 77.21 & 55.67 & 61.02 & 18.93 & 57.25 & 45.15 & 36.62 & 49.71 \\
 & \textsc{Maj-Baseline} & 0.11 & 31.75 & 23.11 & 32.03 & 30.97 & \;\;4.42 & 31.21 & 30.75 & 32.61 & 6.28 \\ \cmidrule(lr){2-12}
 & human upper bound & 0.03 & 93.29 & 90.71 & 81.77 & 82.11 & 30.58 & 78.68 & 88.99 & 79.96 & 95.04\\
  \bottomrule
\end{tabular}
\caption{Performance of our architectures and human upper bound. The \textsuperscript{+} indicate how many other methods (excluding \textsc{Maj-Baseline}) this method significantly outperforms (Mann-Whitney U Test, $P < 0.05$ with Bonferroni correction).}
\label{tab:f1Scores}
\end{table*}

As a reference, we use two baselines: First, a majority baseline, denoted \textsc{Maj-Baseline}, which always predicts the
most frequent $C' \subseteq C$,
resulting in the single label
$I \mhyphen EE$ (continuation EE) for all tokens.
Second, a single-label multi-class classification setup \textsc{Pref-Baseline}, which trains a BiLSTM-CRF to predict a single label (rather than multiple) from $C = (\{B,I\} \times A) \cup \{O\}$. This necessarily ignores any overlaps of epistemic activity segments.
To train this model, we apply the following preference order over epistemic activities derived from the label frequencies in our corpus (see Table~\ref{tab:statistics}) and an importance rating by educational psychologists:
$DC \succ HG \succ EG \succ EE$. The most frequent activity $EE$ receives the lowest preference (i.e.~it is only used as a label if it does not overlap with other activities), while $DC$
is preferred over all other labels due to its importance and to ensure a sufficient amount of training instances. 
Note that the preference order is applied on a segment rather than token level. For example, the first token of the last sentence in Figure~\ref{fig:reasoning_text} receives the label $B \mhyphen DC$, all following tokens before the full stop $I \mhyphen DC$ and the full stop itself the label $O$.

\textbf{Experimental Setup:}
We split our data into 60\% train, 20\% dev, and 20\% test sets, using the same proportion of case scenarios in all splits.
We perform ten runs for each architecture, applying the following parameters for all of them:
one hidden layer of 100 units, variational dropout rates for input and hidden layer of 0.25, and the \emph{nadam} optimizer.
We furthermore use the German \emph{fastText} word embeddings \cite{GraveBGJM2018}. 

%%%%%%%%%%%%%%%%%%%%%%%%%%%%%%%%%%%%%%%%%%%%%%%%%%%%%%%%%%%%%%%%%%
\subsection{Evaluation Metrics}

Multi-label classification tasks are typically evaluated using \emph{hamming loss} \cite{loza14twitterml,SOKOLOVA2009427}. 
This metric quantifies the 
amount of incorrect labels per token, averaged over all tokens: 
\begin{align}
\mathit{HL} = \frac{1}{|\mathcal{X}|} \sum_{x \in \mathcal{X}} \frac{1}{|C|} \sum_{c \in C} \mathbf{xor}(y_{x,c}, \: \hat{y}_{x,c})
\end{align}
where $\mathcal{X}$ is the dataset (set of tokens to be classified),
$y_{x,c}$ is $1$ if token $x$ has label $c$ in the the gold data, and $0$ otherwise, and $\hat{y}_{x,c}$ is $1$ if token $x$ has label $c$ in the prediction, and $0$ otherwise. $\mathbf{xor}(\cdot, \cdot)$ is the usual exclusive-or function.

Given the three challenges associated with the automatic identification of epistemic activities that a machine learning method should be able to tackle, we propose to evaluate each challenge using a separate metric, rather than merely relying on one overall performance score such as $\mathit{HL}$. 
Regarding C1, we measure the segmentation performance in terms of how well the segmentation labels $S$ are predicted for each epistemic activity, thus obtaining a separate performance score for each  $a \in A$:
\begin{align}
M_S(a) = \textit{macro-F1}(C_{a}, \mathcal{X})
\end{align}
where 
$\textit{macro-F1}$ is the macro-averaged harmonic mean between precision and recall across all labels in $C_a$.

The performance of C2 is assessed as performance in predicting the correct epistemic activity, or a combination thereof\footnote{The empty set in $\mathscr{P}(A)$ denotes label $O \in C$. Note that the upper bound of $M_A$ is $62.5$ due to label sets that never occur in the gold data, e.g.~$\{HG,EG,EE,DC\}$.}, for each token. Whether a token constitutes the beginning or continuation of an epistemic activity is disregarded by this metric:
\begin{align}
M_A = \textit{macro-F1}(\mathscr{P}(A), \mathcal{X})
\end{align}

To evaluate C3 -- the ability to predict overlapping segments -- we measure the performance only on those tokens that are associated with at least two labels, i.e.~where for the set of associated labels $C' \subseteq C$ it holds that $|C'| \geq 2$. We denote this set of tokens by $\mathcal{X}_\mathrm{overlap}$ and again obtain a separate performance score for each $a \in A$:
\begin{align}
&M_O(a) = \textit{macro-F1}(C_{a},\mathcal{X}_\mathrm{overlap})
\end{align}

In the next section, we evaluate our different neural architectures using $\mathit{HL}$ and the three challenge-specific metrics to measure how well the architectures address the challenges.

%%%%%%%%%%%%%%%%%%%%%%%%%%%%%%%%%%%%%%%%%%%%%%%%%%%%%%%%%%%%%%%%%%
\subsection{Analysis and Discussion of Results}
Table~\ref{tab:f1Scores} presents our performance results in terms of averages over ten runs.
Since each of the four metrics evaluates different class labels and sets of tokens, a mapping from the outputs of all neural architectures to the respective classes and token sets is applied.
For better interpretability of the scores achieved by the  architectures, Table~\ref{tab:f1Scores} also reports the best score for each metric achieved by some annotator as the human upper bound.

\textbf{Overall Performance. } 
As can be expected given the complexity of this task, the \textsc{Maj-Baseline} does not address any of the three challenges, performing poorly across all performance metrics.
In contrast, \textsc{Pref-Baseline} achieves comparable results to the other neural architectures.

All three neural architectures achieve similar overall performance in terms of the hamming loss $\mathit{HL}$, confirming the necessity for additional performance measures able to assess performance along multiple dimensions.
As there is still a prominent performance gap to the human upper bound, we now analyse each identified challenge in detail to understand which research direction is most promising to improve the current state-of-the-art methods in the future.

\textbf{C1. } 
Regarding the challenge of segmentation and the associated metric $M_S$, 
we again observe that all architectures exhibit similar performance, including our second baseline model \textsc{Pref-Baseline}.
It only exhibits a disadvantage in predicting EE segments, which can be attributed to the `penalisation' of EE in cases where EE overlaps with other epistemic activities (due to the preference order). In such cases, \textsc{Pref-Baseline} is trained to predict one of the other epistemic activities, rather than EE.

Overall, segmenting EE is easiest for the other architectures. One reason for this may be the less skewed distribution of segment labels ($B$: 6\%, $I$: 53\%, $O$: 43\%) as compared to the other epistemic activities, which all exhibit a similarly skewed distribution of $B$: 1\%, $I$: 9\%, and $O$: 90\%. The better performance on EG compared to HG and DC can thus be attributed to easier identifiability of EG, for example due to linguistic cues, rather than the label distribution. 

We observe that the performance trends on segmenting the different epistemic activities as well as the differences in MeD and TeD match the trends in the human upper bound.
Overall, our architectures achieve 78--89\% of the human upper bound on MeD and 70--88\% on TeD, indicating that state-of-the-art methods can perform segmentation reasonably well. 

\textbf{C2. } 
Concerning the distinction of epistemic activities as measured by $M_A$, all architectures achieve less than 66\%\ (MeD) and 71\% (TeD) of the human upper bound, with the 
\textsc{Concat} architecture performing best. Our detailed analysis reveals that this is due to the more reliable prediction for tokens associated with only one epistemic activity compared to the other architectures. 
However, the relatively low performance highlights that the task of distinguishing different epistemic activities is highly challenging.

\begin{figure}[ht]
\centering
\includegraphics[scale = 0.46]{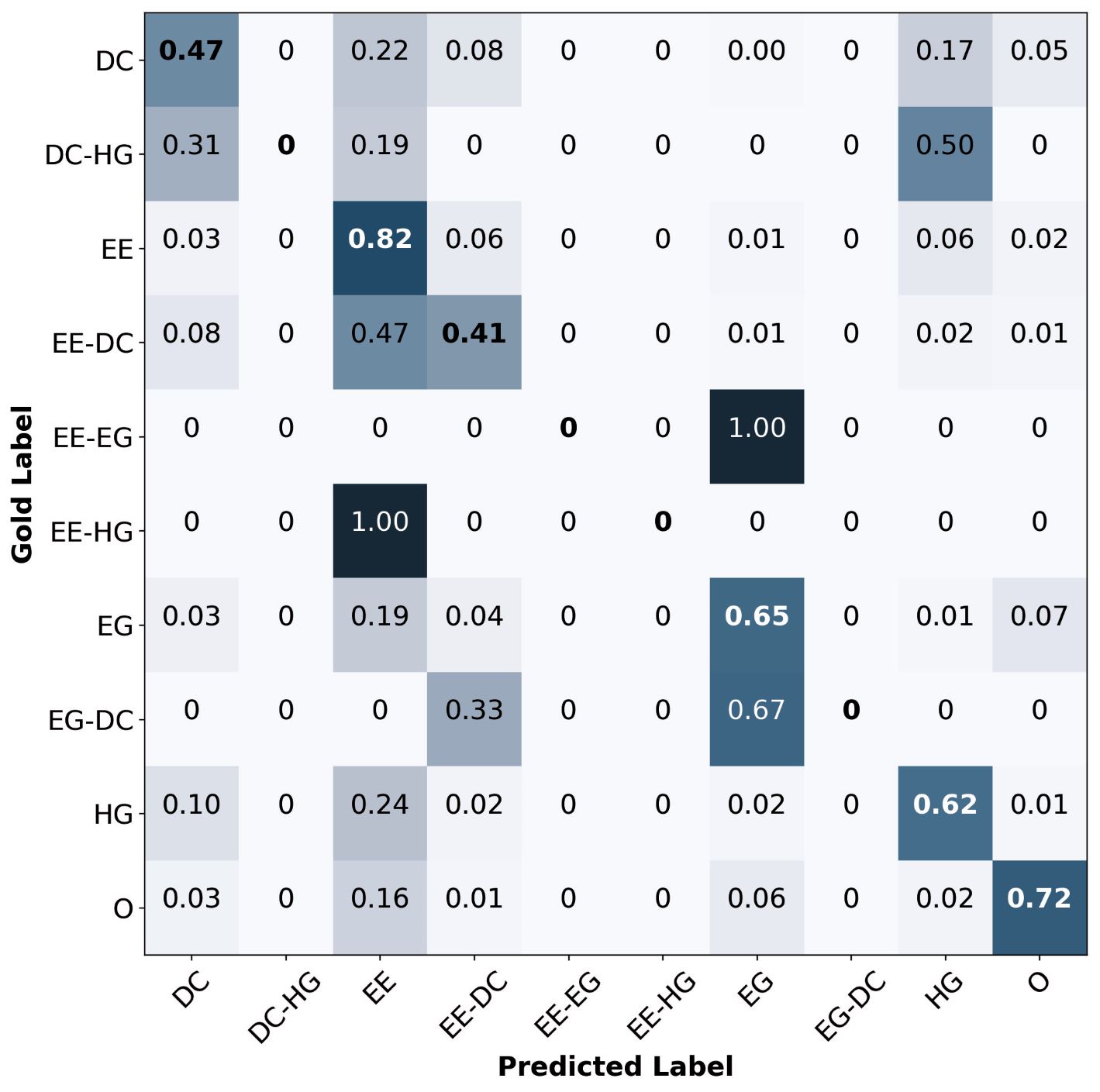}
\caption{Confusion matrix of \textsc{Concat} for MeD.}
\label{fig:confMatrix}
\end{figure}

Figure~\ref{fig:confMatrix} shows an exemplary confusion matrix (of \textsc{Concat} for MeD) regarding the task measured by $M_A$, i.e.~for labels $\mathscr{P}(A)$. Each row indicates the percentage of tokens associated with this label that were (in)correctly classified. Labels that do not occur in the gold data or in predictions are omitted in the figure.
We observe that most confusion occurs due to prediction of the majority label EE, indicating that the skewed distribution of epistemic activities is a problem for the neural architectures.
Furthermore, DC and HG are frequently confused, mirroring the distinction difficulty observed during corpus creation.

\textbf{C3. } 
The evaluation of overlapping segments in terms of $M_O$ shows that our architectures can reasonably identify EE in overlapping segments (87\%/89\% of human upper bound in MeD/TeD), but perform poorly on identifying HG segments involved in overlaps (55\%/46\% of human upper bound in MeD/TeD). This may be due to the small amount of overlaps involving HG (see Table~\ref{tab:statistics}). However, the equally infrequent overlaps involving EG can be predicted more reliably, whereas the frequently occurring overlaps with DC achieve poor results. Thus, the poor performance on overlaps is not only due to the skewed label distribution, but also to the inherent difficulty of overlapping epistemic activities, which is highly challenging for
state-of-the-art methods.

The confusion matrix (Figure~\ref{fig:confMatrix}) sheds further light on the prediction of overlaps:
For overlaps of two epistemic activities, our architectures often identify only one of them. For example, $EE\mhyphen DC$ is mostly wrongly predicted to be only $EE$ or only $DC$. Note that \textsc{Concat} only predicts overlaps between epistemic activities that also occur in the training data. In contrast, \textsc{Separate} also predicts other overlaps, even overlaps between three epistemic activities.

%%%%%%%%%%%%%%%%%%%%%%%%%%%%%%%%%%%%%%%%%%%%%%%%%%%%%%%%%%%%%%%%%%
%%%%%%%%%%%%%%%%%%%%%%%%%%%%%%%%%%%%%%%%%%%%%%%%%%%%%%%%%%%%%%%%%%
\section{Conclusion and Discussion}

We presented a novel corpus of diagnostic reasoning texts written by medical students and pre-service teachers, which are annotated with epistemic activities. We identified three intrinsic characteristics of our corpus, which constitute \emph{challenges} for AI methods aiming to automatically identify epistemic activities.
To measure the performance of AI systems on each of these challenges, we propose a separate performance metric for each challenge. These provide an evaluation framework for future research. 

Indeed, we show that state-of-the-art recurrent neural network architectures are unable to satisfactorily tackle two of the three challenges, namely the accurate distinction of different epistemic activities and the correct identification of overlapping epistemic activity segments.
This leads to two conclusions: On the one hand, it is crucial to evaluate AI systems along different task-specific characteristics, instead of simply using common overall metrics, such as the overall F1 score or hamming loss. On the other hand, future work needs to focus on the development of systems specifically designed to tackle the challenges of our novel corpus. Such systems will also prove useful for other multi-label sequence labeling tasks.

We expect that our work will have a large impact for educational psychologists as it is a crucial step towards automatically generating feedback on students' reasoning at a large scale. 
As a next step, we will investigate new approaches to better solve the three challenges of epistemic activity identification and extend our task to the automatic assessment and feedback generation for diagnostic reasoning.

%%%%%%%%%%%%%%%%%%%%%%%%%%%%%%%%%%%%%%%%%%%%%%%%%%%%%%%%%%%%%%%%%%
%%%%%%%%%%%%%%%%%%%%%%%%%%%%%%%%%%%%%%%%%%%%%%%%%%%%%%%%%%%%%%%%%%
\section*{Acknowledgments}
This work was supported by the German Federal Ministry of Education
and Research (BMBF) under the reference 16DHL1040 (FAMULUS).

\bibliography{edas_minimal}
\bibliographystyle{aaai}

\end{document}